\documentclass[journal]{IEEEtran}
\usepackage{array}
\usepackage{url}
\usepackage{cite}
\usepackage{booktabs}
\usepackage{amsfonts}
\usepackage{multirow, subfigure, graphicx, caption,color}
\usepackage{tabularx}
\usepackage{epstopdf}
\usepackage{amsmath}
\usepackage[ruled]{algorithm2e}
\usepackage{bm}
\usepackage[switch]{lineno}
\usepackage{makecell}
\usepackage[numbers]{natbib}
\usepackage{amssymb}
\usepackage{arydshln}
\usepackage{algorithmic}
\newcommand{\tabincell}[2]{
\begin{tabular}{@{}#1@{}}#2\end{tabular}
}

\newcolumntype{L}[1]{>{\raggedright\let\newline\\\arraybackslash\hspace{0pt}}m{#1}}
\newcolumntype{C}[1]{>{\centering\let\newline\\\arraybackslash\hspace{0pt}}m{#1}}
\newcolumntype{R}[1]{>{\raggedleft\let\newline\\\arraybackslash\hspace{0pt}}m{#1}}
\newcommand\Tstrut{\rule{0pt}{2.6ex}}         
\begin{document}

\title{Weakly Supervised Patch Label Inference Networks for Efficient Pavement Distress Detection and Recognition in the Wild}

%

\author{Sheng Huang,~\IEEEmembership{Member,~IEEE,}
        Wenhao Tang, Guixin Huang, Luwen Huangfu,
        and~Dan Yang
\thanks{This work was supported in part by the National Natural Science Foundation of China under Grant 62176030, and in part by the Natural Science Foundation of Chongqing under Grant cstc2021jcyj-msxmX0568. ( S. Huang is the corresponding author.)}
\thanks{S. Huang is with Ministry of Education Key Laboratory of Dependable Service Computing in Cyber Physical Society, Chongqing, 400044, P.R.China,
S. Huang, W. Tang, G. Huang, and D. Yang are with the School of Big Data and Software Engineering, Chongqing University, Chonqqing, 400044 P.R.China,(email:\{huangsheng, whtang, huangguixin, dyang\}@cqu.edu.cn),}
\thanks{L. Huangfu is with Fowler College of Business and also center for Human Dynamics in the Mobile Age, San Diego State University, San Diego, California, 92182, USA, (email:lhuangfu@sdsu.edu)}}

%
%

\markboth{Published on IEEE TRANSACTIONS ON INTELLIGENT TRANSPORTATION SYSTEMS}%
{Shell \MakeLowercase{\textit{et al.}}: Bare Demo of IEEEtran.cls for IEEE Journals}
%



%
\maketitle
%
%

%

%
\begin{abstract}
Automatic image-based pavement distress detection and recognition are vital for pavement maintenance and management. However, existing deep learning-based methods largely omit the specific characteristics of pavement images, such as high image resolution and low distress area ratio, and are not end-to-end trainable. In this paper, we present a series of simple yet effective end-to-end deep learning approaches named Weakly Supervised Patch Label Inference Networks (WSPLIN) for efficiently addressing these tasks under various application settings. WSPLIN transforms the fully supervised pavement image classification problem into a weakly supervised pavement patch classification problem for solutions. Specifically, WSPLIN first divides the pavement image under different scales into patches with different collection strategies and then employs a Patch Label Inference Network (PLIN) to infer the labels of these patches to fully exploit the resolution and scale information. Notably, we design a patch label sparsity constraint based on the prior knowledge of distress distribution and leverage the Comprehensive Decision Network (CDN) to guide the training of PLIN in a weakly supervised way. Therefore, the patch labels produced by PLIN provide interpretable intermediate information, such as the rough location and the type of distress. We evaluate our method on a large-scale bituminous pavement distress dataset named CQU-BPDD and the augmented Crack500 (Crack500-PDD) dataset, which is a newly constructed pavement distress detection dataset augmented from the Crack500. Extensive results demonstrate the superiority of our method over baselines in both performance and efficiency. The source codes of WSPLIN are released on https://github.com/DearCaat/wsplin.
\end{abstract}

\section{Introduction}
\label{sec:intro}
With the rapid growth of society and the modern logistics industry, road infrastructure has been greatly increased in today's world. There are a total of more than 64,000,000 kilometers of roads in the world~\cite{length}, which leads to massive operational requirements for pavement maintenance. The pavement inspection is one of the key steps~\cite{c1}. Generally speaking, the cameras are often utilized as pavement inspection equipment due to their low cost and the powerful data representational ability of images. Therefore, the pavement inspection task is often translated into a pavement distress analysis task based on the acquired pavement images, and then this task is accomplished manually by proficient workers. Clearly, such an operation consumes plenty of time and labor resources due to an enormous amount of pavement images produced daily~\cite{c2}.
Therefore, automating the pavement distress analysis is critical in improving efficiency, reducing cost, and avoiding labeling errors in manual pavement inspection.

\begin{figure}[t]
    \centering
    \includegraphics[scale=0.35]{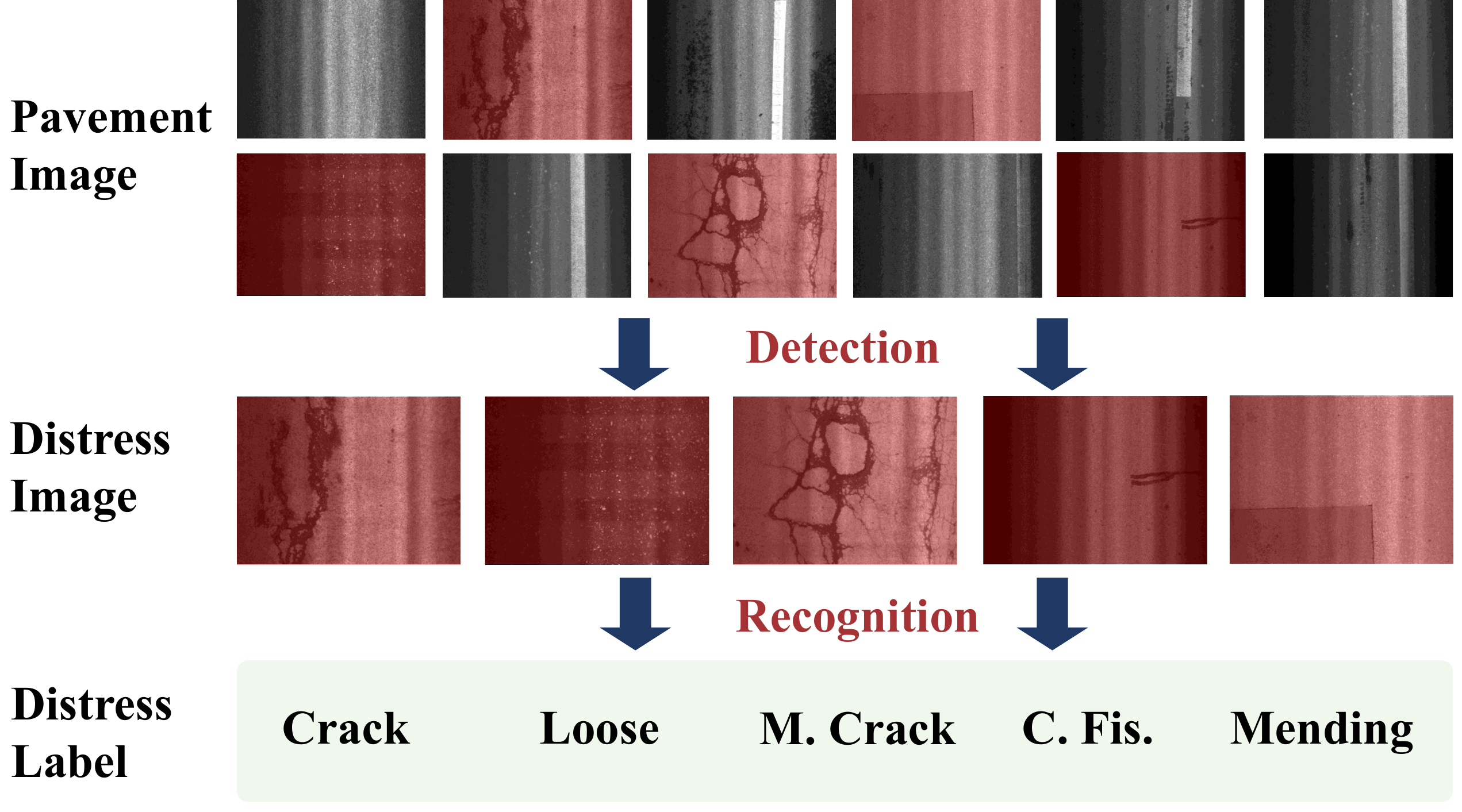}
    \caption{The illustrations of pavement distress detection (finding out the distress images) and recognition (recognizing the pavement distresses).}
    \label{fig.firstpage}
    \vspace{-0.5cm}
\end{figure}

Pavement distress detection and recognition are two fundamental tasks for pavement distress analysis, which aim at filtering out the distressed pavements and classifying the distressed pavements into specific categories, respectively.
This task is distinct from crack detection, which focuses on locating and segmenting cracks on various surfaces. Pavement distress classification (PDC)aims at distinguishing various types of pavement distress, which may or may not be cracks. Both of these two tasks present their own challenges. It is particularly difficult challenging to accurately segment the complex cracks in crack detection due to the irregular structures of these complex cracks. In real-world scenarios, most of the collected pavement images often either do not contain any distress or only have very small, subtle distress areas, as shown in Figure~\ref{fig.firstpage}. This situation makes the PDC model more difficult to distinguish the distressed pavements from the normal ones, and the complex cracks are the key features for benefiting the classification. Clearly, small cracks or diseased areas are more likely to be challenging factors in PDC. In recent decades, many classical approaches have been applied to PDC.

The first group is to utilize image processing, hand-craft features, and conventional classifiers to recognize pavement distress~\cite{c3,c27,c28,c29,c30}. For example, Zhou et al.~\cite{c29} developed a two-step method that conducts the wavelet transform followed by a random transform to classify pavement distress. Sun et al.~\cite{c3} proposed a crack classification method based on topological properties and chain code. The main drawback of these methods is that they often optimize the feature extraction and classification step separately or even do not involve any learning process, which leads to poor performance. Moreover, it usually needs plenty of sophisticated image pre-processing.

Inspired by the advance of deep learning approaches, it is more and more popular to apply different deep learning-based visual learning models for pavement distress detection and recognition~\cite{c8,c9,c10,c11,c16}. For example, Laha et al.~\cite{rddretina} detected road damage with RetinaNet~\cite{retinanet}. Compared to conventional approaches, deep learning-based approaches often achieve better performance. However, most of these approaches only regard the pavement distress detection or recognition problem as common object detection or image classification problem and directly apply the classical deep learning approaches. They seldomly paid attention to the specific characteristics of pavement images, such as the high image resolution, the low distress area ratio.

In order to address the aforementioned issue, IOPLIN~\cite{c16} employs a Patch Label Inference Network (PLIN) to infer the labels of patches from pavement images, taking advantage of the high-resolution image information. It elaborates an Expectation-Maximization-based strategy and a Patch Label Distillation strategy to iteratively train the network with only image labels, and has achieved remarkable detection performances. This approach represents a brand-new and innovative solution that emphasizes local information extraction and transforms the problem into a weakly supervised patch classification problem for classifying pavement images. However, its model optimization procedure is complex and time-consuming. Moreover, this model is also not end-to-end trainable and cannot be further extended to address the pavement distress recognition issue.

To address the aforementioned issues, we present a novel pavement image classification framework named Weakly Supervised Patch Label Inference Network (WSPLIN)~\cite{wsplin} for both pavement distress detection and recognition. Similar to the IOPLIN, WSPLIN also accomplishes the pavement image classification via inferring the labels of patches from the pavement images with Patch Label Inference Networks (PLIN).
Thus, WSPLIN inherits the merits of IOPLIN, such as the better image resolution information exploitation and result interpretability, but also suffers from the obstacle of training PLIN only with image labels. Compared to IOPLIN, WSPLIN solves this issue via introducing a more concise end-to-end patch-level weakly supervised learning framework. Such a framework endows WSPLIN with better efficiency and greater flexibility, enabling the pavement distress recognition application.

In WSPLIN, the pavement image is divided into patches with different patch collection strategies under different scales for exploiting both global and local information. Then, a CNN is implemented as PLIN for inferring the labels of patches with a sparsity constraint. Finally, the patch label inference results are fed into a Comprehensive Decision Network (CDN) for completing the classification. We integrate PLIN and CDN as an end-to-end deep learning model. In such a manner, the PLIN can be optimized by the guidance of CDN and the patch label sparsity constraint in a cleaner and more efficient fashion. Moreover, three different strategies, namely Sliding Window (SW), Image Pyramid (IP), and Sparse Sampling (SS), are adopted for collecting patches from images. We name these corresponding WSPLIN versions, WSPLIN-SW, WSPLIN-IP, and WSPLIN-SS, respectively. As same as IOPLIN, WSPLIN-SW has not considered any scale information during patch collection. It can be deemed as a naive version of WSPLIN. Different from WSPLIN-SW, WSPLIN-IP incorporates the scale information via dividing images into patches from coarse to fine based on an image pyramid. It is the default version of WSPLIN. WSPLIN-SS conducts a sparse patch sampling to collect only a few patches from the image pyramid to improve the efficiency of WSPLIN. It can be seen as the lightweight version of WSPLIN. We evaluate WSPLIN on a large-scale pavement image dataset named CQU-BPDD~\cite{c16} and a new pavement image classification dataset constructed from Crack500 under different settings, including distress detection, one-stage recognition, and two-stage recognition. The experimental results show that WSPLIN outperforms extensive baselines and demonstrates prominent advantages over IOPLIN in both efficiency and performance.

The main contributions of our work are summarized as follows:
\begin{itemize}
    \item We propose a novel end-to-end patch-level weakly supervised deep learning model named WSPLIN for addressing both pavement distress detection and recognition issues. WSPLIN transforms the fully supervised pavement image classification task into a weakly supervised patch-level label inference issue for the solution. It not only inherits the merits of IOPLIN, but also enjoys faster training speed, better classification performance, and wider application scenarios over IOPLIN.
    \item Different from IOPLIN and the conventional CNN-based image classification methods, we introduce image pyramid to WSPLIN-IP for exploiting scale information. Moreover, we design a sparse patch sampling strategy in the image pyramid for further speeding up WSPLIN. The model training time of this faster WSPLIN version (WSPLIN-SS) is only one-fourth of the training time of IOPLIN, while they share similar performance in pavement distress detection.
    \item We introduce a simple constraint named patch-label sparsity constraint (PLSC) to incorporate the prior knowledge of the pavement distresses together with the Comprehensive Decision Network (CDN) for further boosting the patch-label inference process. The patch labels produced by PLIN enable providing some interpretable intermediate information, such as the rough location and the type of distress.
    \item We empirically evaluate our model compared with some tailored methods and several state-of-the-art deep learning-based image classification approaches as baselines in pavement distress detection and recognition. Extensive results demonstrate that our proposed method outperforms them in all tasks under different settings on a large-scale bituminous pavement distress dataset named CQU-BPDD and a newly augmented pavement distress detection dataset constructed from Crack500.
\end{itemize}

\section{Related Work}
\label{sec:rela_work}
\subsection{Image-based Pavement Distress Analysis}
The traditional pavement distress analysis approaches mainly include filter-based methods and hand-crafted feature-based classical classifiers. For example, in~\cite{zhou2006wavelet}, wavelet transform is used to decompose a pavement image into different-frequency subbands. Hu et al.~\cite{hu2010novel} propose a novel Local Binary Pattern (LBP) based operator for pavement crack detection. In~\cite{shi2016automatic}, a random structured forest named CrackForest, which is combined with the integral channel features, is proposed for automatic road crack detection. Kapela et al.~\cite{kapela2015asphalt} propose a crack recognition system based on the Histograms of Oriented Gradients (HOG). Pan et al.~\cite{pan2017object} use the four popular supervised learning algorithms (KNN, SVM, ANN, RF) to discern pavement damages. However, the traditional methods usually have weak performance owing to numerous artificial design factors, and separate optimization procedures, and they cannot be adapted to a large number of data currently.

Inspired by the recent remarkable successes of deep learning in extensive applications, simple and efficient convolutional neural networks (CNN) based pavement distress analysis methods have gradually become mainstream in recent years. In general, these methods can be divided into three parts according to the task objective: pavement distress segmentation~\cite{c11,crack500,yang2019feature,zou2018deepcrack}, pavement distress location~\cite{ibragimov2020automated,ZHU2022103991}, and pavement distress classification~\cite{c16,few_shot,dong2021automatic}. Among them, pixel-based pavement distress segmentation is a hot research field. Zhang et al.~\cite{crack500} leverage CNN to classify the image patch for segmenting pavement distress. In~\cite{c11}, a CNN is used to learn the structure of the cracks from raw images, then the segmentation result is generated by the obtained structure information. Based on the fully conventional network (FCN), Yang et al.~\cite{yang2019feature} fuse multiscale features from top-to-down for pavement crack segmentation. In DeepCrack~\cite{zou2018deepcrack}, multiscale deep convolutional features learned at hierarchical convolutional stages are fused together to capture the line structures. For distress localization, Ibragimov et al.~\cite{ibragimov2020automated} propose a method for localizing signs of pavement distress based on a faster region-based conventional neural network. Zhu et al.~\cite{ZHU2022103991} compare the performance of three state-of-the-art object-detection algorithms on an Unmanned aerial vehicles(UAV) pavement image dataset, which includes six types of distress. Because pavement distress annotation requires professional knowledge and a large amount of time, the datasets used in the above methods are low-resolution and small-scale. However,  it remains to be determined whether models derived from small-scale datasets can be applied to real-world practice.

For pavement distress classification, Dong et al.~\cite{few_shot} propose a metric-learning based method for multi-target few-shot pavement distress classification on the dataset, which includes ten different kinds of distress. In~\cite{dong2021automatic}, discriminative super-features constructed by the multi-level context information from the CNN are used to determine whether there is distress in the pavement image and recognize the type of distress. All of these methods do a good job of classification on the dataset they use, which is small and only contains distressed images, and on which the test accuracy even achieves 100\%~\cite{dong2021automatic}. There have been few works to systematically evaluate the model's performance on a difficult large-scale multi-type dataset. Moreover, these approaches only regard the pavement distress detection or recognition problem as a common image classification problem and directly apply the classical deep learning approaches. In~\cite{c16}, patch-based weakly learning model IOPLIN and large-scale distress datasets CQU-BPDD are proposed to solve these problems. However, the main drawback of IOPLIN is that the patch label inference strategy based on the pseudo label makes IOPLIN incompatible with pavement recognition, and its optimization process is quite complex and time-consuming. Our approach takes inspiration from IOPLIN but operates with different patch inference strategy and uses more effective and hierarchical patch collection strategies. Besides, Li et al.~\cite{stn} leverage Spatial Transformer Network~\cite{jaderberg2015spatial} to automatically recognize different types of defects and measure the coverage percentage on the ship surface, which introduces a possible solution for pavement image classification.

\vspace{-0.2cm}
\subsection{Deep Learning-based Image Classification}
In recent years, due to the popularity of the ImageNet Large Scale Visual Recognition Challenge (ILSVRC)~\cite{ilsvrc}, many computer vision algorithms based on deep learning have emerged. Among them, a series of convolutional neural networks (CNNs) play a leading role in the field of image classification. For example, AlexNet~\cite{alexNet} first applies the structure of convolutional neural networks to large-scale image classification datasets. Simonyan et al. first propose the deep and large-scale convolutional neural network VGGNet~\cite{c17} (i.e., the VGG19 model has 19 layers and more than 130 million parameters, while the previous convolutional neural network has less than 10 layers and millions to tens of millions of parameters). In InceptionNet~\cite{c18}, $1 \times 1$ convolution kernel application is proposed for the first time. He et al.~\cite{c19} propose a residual structure and network extension strategy to construct a network family for the first time. Zoph et al.~\cite{mobilenets} bring CNN into the embedded mobile terminal and propose MobileNet, specially designed for low computing power and low memory computing platform. Then based on MobileNet and Neural architecture search (NAS)~\cite{nas}, Tan et al. propose an efficient CNN, dubbed EfficientNet~\cite{c21}. In the past year, inspired by the field of Natural Language Processing (NLP), Dosovitskiya et al.~\cite{vit} propose a visual classification model based on Transformer~\cite{attention}, named as ViT.

However, it is very challenging to employ the general classification models based on CNN and Transformer for addressing pavement distress analysis issues directly. This is mainly because pavement images have many specific characteristics, such as high image resolution, low distress area ratio, and uneven illumination, compared with object-centric natural images. Our approach intends to employ a patch collection strategy in an image pyramid to incorporate both local and multi-scale information for pavement distress analysis.


\begin{figure*}[t]
    \centering
    \includegraphics[scale=0.6]{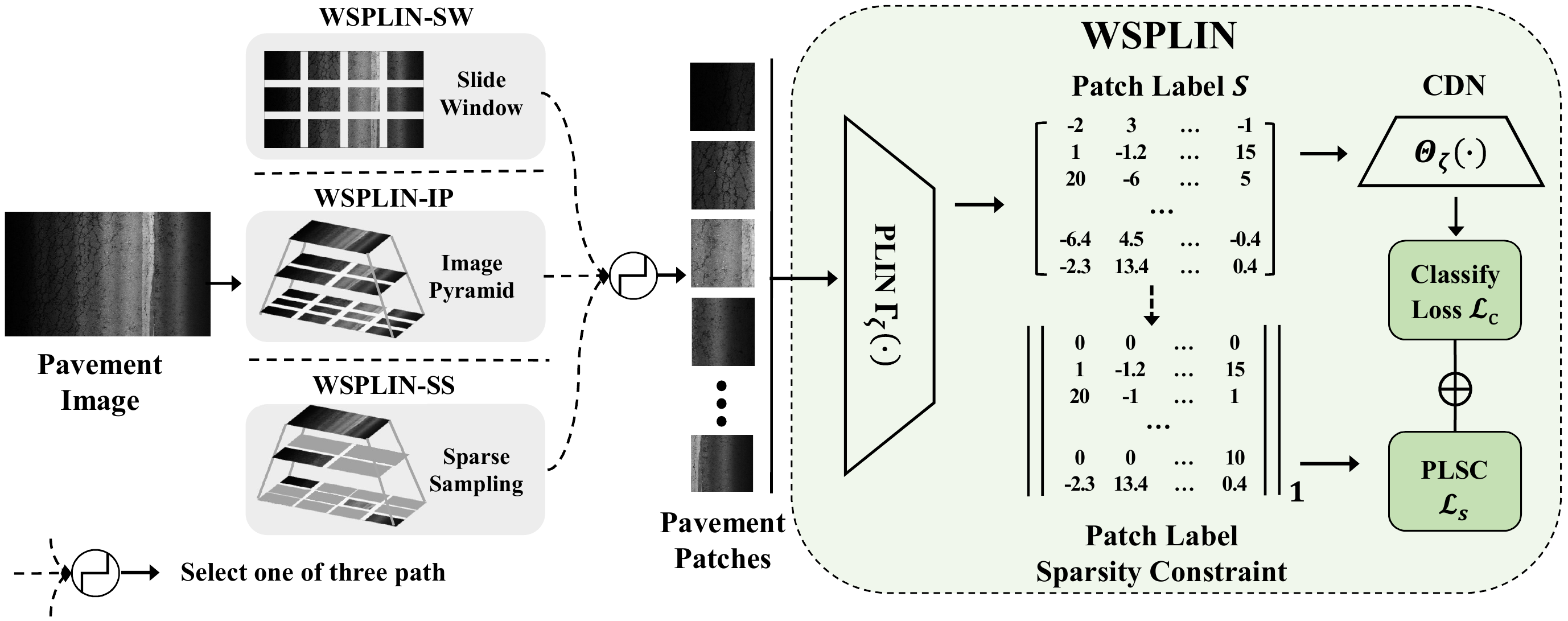}
    \caption{The overview of Weakly Supervised Patch Label Inference Network (WSPLIN). WSPLIN transforms the fully supervised pavement image classification problem into a weakly supervised pavement patch classification problem for solutions. Specifically, WSPLIN employs different patch collection strategies to divide the pavement images into patches under different scales. Then, a Patch Label Inference Network (PLIN) is leveraged to produce the pseudo labels of patches. Finally, a Comprehensive Decision Network (CDN) is established for yielding the final pavement image label based on the produced patch labels.}
    \label{fig.network}
    \vspace{-0.5cm}
\end{figure*}

\section{Methodology}
\label{sec:format}
The network architecture of the Weakly Supervised Patch Label Inference Network (WSPLIN) is shown in Figure~\ref{fig.network}. In this section, we first introduce the problem formulation and overview of WSPLIN in section~\ref{sec:formulation}. Then, we introduce the involved patch collection strategies in section~\ref{sec:pc}. After that, the core modules of WSPLIN, Patch Label Inference Network (PLIN), and Comprehensive Decision Network (CDN) are detailed in section~\ref{sec:plin} and section~\ref{sec:cdn} respectively. Finally, we will show how to apply WSPLIN to detect and recognize the pavement distress in section~\ref{sec:pavement DR}.

\vspace{-0.2cm}
\subsection{Problem Formulation and Overview}
\label{sec:formulation}
Both pavement distress detection and recognition can be deemed as image classification tasks from the perspective of computer vision. Let $X=\{x_1,\cdots,x_n\}$ and $Y=\{y_1,\cdots,y_n\}$ be the collection of pavement images and their pavement labels, respectively. $y_i$ is a $C$-dimensional one-hot vector where $C$ is the number of categories and $y_{ij}$ indicates the $j$-th element of $y_i$. In the detection case, such a classification task is a binary image classification issue (distressed or normal) where $C=2$. In the recognition case, this classification task is a multi-class image classification problem where $C>2$. In a pavement label $y$, if the $j$-th element is the only nonzero element, it indicates that the corresponding pavement image belongs to the $j$-th category.
The pavement distress detection or recognition is to learn a classifier $F_{det}(\cdot)$ or $F_{reg}(\cdot)$ can label the pavement image correctly, $y_i\leftarrow F(x_i)$.

There are two strategies for accomplishing the pavement distress recognition task. One is the two-stage recognition flow path, and the other is the one-stage recognition flow path. The two-stage recognition is to identify distressed images first via pavement distress detection and then apply the pavement distress recognition to further classify each distressed image into a specific type of pavement distress. The one-stage recognition directly considers the normal case as an additional category in the recognition procedure. Therefore, the pavement distress detection and recognition tasks are jointly tackled with one image classification model.

Similar to Iteratively Optimized Patch Label Inference Network (IOPLIN), WSPLIN is a patch-based pavement image classification method whose main obstacle is to train Patch Label Inference Network (PLIN) only with the image label. WSPLIN introduces an additional module named Comprehensive Decision Network (CDN) to guide the optimization of PLINs in an end-to-end weakly supervised learning manner. The flow path of WSPLIN is very concise. In WSPLIN, the pavement image is divided into several patches first, and then PLIN is used to infer the labels of these patches. Finally, the inferred labels are fed into CDN to yield the final pavement label. WSPLIN has two core modules, namely PLIN and CDN, whose corresponding mapping functions are $\Gamma_\zeta(\cdot)$ and $\Theta_\xi(\cdot)$, respectively.

\vspace{-0.2cm}
\subsection{Patches Collection}
\label{sec:pc}

We adopt three different patch collection strategies for producing patches. They are Slide Window (SW), Image Pyramid (IP), and Sparse Sampling (SS). The first strategy is also adopted by IOPLIN. WSPLIN uses the second strategy to exploit image information from different scales fully. The third strategy is newly designed by us based on the IP strategy for speeding up WSPLIN via reducing the patch amount for training.

\textbf{Slide Window}:
The pavement image is simply divided into a series of uniform scale patches following a non-overlapping strategy. We adopt $300\times300$ as the sliding window size with 300 sliding stride. The patch collection can be denoted as $P_i= \tau(x_i) = \{p^i_1, ..., p^i_m\}, m = 12$ where $\tau(\cdot)$ is our patch extraction operation.

\textbf{Image Pyramid}:
The slide window strategy does not consider the scale information. So we resize the pavement image into three resolutions, $300 \times 300$, $600\times 600$ and $1200 \times 900$ (the original size), to construct a three-layer image pyramid from top to down, and then employ sliding window method for dividing the image into patches. The patch collection can be denoted as $P_i= \{\tau(x_i^l)\}_{l\in\{0,1,2\}} = \{p^i_1, ..., p^i_m\}, m=\sum_l m_l = 17$ where $x_i^0=x_i$ and $l$ indicates the layer ID. Similar to the slide window strategy, we also apply $300\times300$ as the sliding window size with 300 sliding strides in the image pyramid. Therefore, $m_0=12$, $m_1=4$, and $m_2=1$.

%
\textbf{Sparse Sampling}:
The patch number determines the scale of training data, and the patches in the same image pyramid also contain some redundant information in the scale space. Therefore, we can sample some patches for each image to reduce the training burden and speed up the model. More specifically, let $\alpha$ be the sparse sample ratio to control
the number of sampled patches for each layer, $n_l=\left \lceil m_{l}\times \left ( 1-\alpha  \right ) \right \rceil$ where $\left \lceil  \cdot \right \rceil$ returns the smallest integer that is greater than or equal to the input. We design a simple strategy for sampling patches in each layer. In this strategy, the sampled patches of all three layers should cover all scales while maximizing spatial coverage. The optimal patch sparse sampling strategy is mathematically denoted as follows,
\begin{equation}
\setlength\abovedisplayskip{1pt}
\setlength\belowdisplayskip{1pt}
   \{\hat{\mathcal{C}}_l\}_{l=0,1,2}\leftarrow\arg\underset{\{\mathcal{C}_l\}_{l=0,1,2}}\max { \text{Vol}(\bigcup_{l=0}^2\bigcup_{t\in \mathcal{C}_l} p_t)}, \quad \text{s.t.} |\mathcal{C}_l|=n_l,
\end{equation}
where $\text{Vol}(\cdot)$ returns the volume of the given set and $\mathcal{C}_l$ denotes an index subset to patches in $l$-th layer. Since the solutions to the above problem are limited, we can use the enumeration method to address this issue efficiently when $n_l$ is fixed. In this paper, we empirically set $\alpha = 0.25$. In such a manner, $n_0=1$, $n_1=1$ and $n_2=3$.

\emph{For distinguishing different versions of WSPLIN, WSPLIN-SW, WSPLIN-IP, and WSPLIN-SS indicate the versions that use sliding window, image pyramid, and sparse sampling patch collection strategies, respectively. The default version of WSPLIN is WSPLIN-IP, since WSPLIN-IP is the best-performing version. WSPLIN-SS is the lightweight version of WSPLIN, which takes a good trade-off between performance and efficiency due to its special design. WSPLIN-SW is an ideal choice in the case that the distresses have not suffered from the high diversity in scale.}

\vspace{-0.2cm}
\subsection{Patch Label Inference Network}
\label{sec:plin}
Similar to IOPLIN~\cite{c16}, we adopt EfficientNet-B3~\cite{c21} as our Patch Label Inference Network (PLIN) due to its good trade-off between performance and efficiency. We denote $\Gamma_\zeta(\cdot)$ as the mapping function of PLIN. The patch label inference procedure is denoted as,
\begin{equation}
S_i = \Gamma_\zeta(\tau(x_i)),
\end{equation}
where $S_i$ is an $m\times C$-dimensional matrix in which every column encodes the label inference confidences of patches. Such confidences are expected to be zero if the patch does not contain the distress and reflect the possibility of the certain distress that the corresponding patch has. Please note, there is no supervised information on patches, so all these labels are randomly produced just via forward propagation. We need to leverage the follow-up comprehensive decision network to guide the PLIN to generate reasonable patch labels with image-level labels in a weakly supervised manner. We will introduce such a procedure later.

\noindent\textbf{Patch Label Sparsity Constraint (PLSC):} Since most pavement distress areas are actually only a tiny fraction of images, most patches in a pavement image should not have any distress, and the model should output low responses to these. In such a manner, the label confidence matrix $S_i$ should only have very limited nonzero elements, which indicates that $S_i$ should be sparse. Thus, we introduce an $L_1$-norm constraint to the label confidence matrices of the distressed training samples,
\begin{equation}
    \mathcal{L}_s=\sum_{i \in \{i|y_i\neq y_{\text{normal}}\}}||S_i||_1,
\end{equation}
where $y_{normal}$ is the label of the normal pavement image. We introduce this constraint only to the distressed samples, since there should be no nonzero element in the label confidence matrices of the normal samples, $||S_i||_1=0$

\vspace{-0.2cm}
\subsection{Comprehensive Decision Network}
\label{sec:cdn}
We establish a Comprehensive Decision Network (CDN) for accomplishing the final pavement image classification based on the aforementioned patch label results. CDN consists of four layers where the first two layers are all the $m\times C$ fully connection layers followed by a ReLU, and Dropout layer, the third layer is also an $m\times C$ fully connection layer, and the size of output fully connection layer is $C$. Here, $C$ is the number of categories. Let $\Theta_\xi(\cdot)$ be the mapping function of CDN, then the predicted pavement distress label $\hat{y}_i$ can be obtained by,
\begin{equation}
    \hat{y}_i = \Theta_\xi(\Gamma_\zeta(\tau(x_i)))=\Theta_\xi(S_i).
\end{equation}
We use the cross-entropy to measure the discrepancy between the predicted label and ground-truth and denote it as the classification loss $\mathcal{L}_c$,
\begin{equation}
    \label{eq:cel}
    \mathcal{L}_c = -\frac{1}{n}\sum_{i=1}^n\sum_{j=1}^C y_{ij}\log{\hat{y}_{ij}}.
\end{equation}
Finally, the optimal WSPLIN model is learned by minimizing the following loss,
\begin{equation}
    (\xi,\zeta)  \leftarrow\arg\underset{\hat{\xi},\hat{\zeta}}\min~~\mathcal{L}_{total}:=\mathcal{L}_c+\lambda\mathcal{L}_s,
\end{equation}
where $\lambda$ is a positive parameter for reconciling the classification loss and the sparsity constraint.

WSPLIN is an end-to-end deep learning framework that uses back-propagation to compute the loss deviation and update the parameters layer by layer. In WSPLIN, CDN requires the patch label results produced by PLIN that should be useful for the final classification. The patch label sparsity constraint forces WSPLIN to highlight only a few of the most crucial patches for participating in the final decision. Clearly, these highlighted patches should be distressed ones, and their inferred patch label results should be nonzero since only the distressed patches can provide helpful information for the final detection and recognition. In such a manner, CDN essentially guides the training of PLIN in a weakly supervised manner.

\vspace{-0.2cm}
\subsection{Pavement Distress Detection and Recognition}
\label{sec:pavement DR}

\textbf{Detection and One-Stage Recognition:}
The pavement detection and one-stage pavement distress detection can be deemed as a one-stage pavement image classification problem. To tackle these tasks, we can train our model as a pavement image classifier. Once the model is trained, the pavement image $x$ can be divided into patches with different patch collection strategies, which are fed into WSPLIN for yielding the final classification,
\begin{equation}
    y = F(x) = \Theta_\xi(\Gamma_\zeta(\tau(x))),
\end{equation}
where $F \in \{F_{det}, F_{reg}\}$ and the predicted category should be corresponding to the maximum element of $y$.

\textbf{Two-Stage Recognition:}
The two-stage recognition has two stages to accomplish pavement distress recognition. The first stage is to train our model as a pavement distress detector $F_{det}$ for filtering out the normal samples and finding the distressed samples. The second stage is to train our model as a multi-class pavement image classifier $F_{reg}$ for completing the final distress recognition,
\begin{equation}
    y_i = F_{reg}(x_i) = \Theta_\xi(\Gamma_\zeta(\tau(x_i))),
\end{equation}
where $x_i \in \{x|F_{det}(x)\neq y_{\text{normal}}\}$ and the maximum element of $y_i$ reflects the specific pavement distress category of the distressed pavement image $x_i$.

\section{Experiments and Results}

\subsection{Dataset and Setup}
\subsubsection{Pavement Image Classification Tasks}
We test our method on pavement distress detection and recognition tasks under four application settings. The first one is the one-stage recognition (\textbf{I-REC}), which tackles the pavement distress detection and recognition tasks jointly. In this setting, all samples (including both the distressed and normal ones) and their fine-grained category label are available for training and testing the model. Moreover, both the detection and recognition performances can be evaluated under this setting. The second one is the one-stage detection (\textbf{I-DET}), which is the conventional detection fashion. In this setting, all samples (including both the distressed and normal ones) are involved, but only the binary coarse-grained category label (distressed or normal) is available. The other two settings are all from the two-stage recognition scenario. One is the ideal second-stage recognition \textbf{II-REC(i)}, which assumes all distressed samples are ideally detected via the first-stage detection. In this setting, the recognition models are only evaluated with distressed pavement images. The last setting is the normal second-stage recognition \textbf{II-REC(n)}. The training stage of \textbf{II-REC(i)} and \textbf{II-REC(n)} are identical. But their testing stages are different. In \textbf{II-REC(n)}, the recognition models are only evaluated on the images detected by the detection model trained in \textbf{I-DET}. In such a manner, the recognition error under this setting is the errors accumulated by both the first-stage detection and the second-stage recognition, since some distressed images may be incorrectly filtered out while some normal images may be incorrectly classified as distressed ones by the detector in the recognition testing stage under \textbf{II-REC(n)}. The results in \textbf{II-REC(n)} can reflect the comprehensive performances of two-stage recognition.

\begin{figure}[t]
    \centering
    \includegraphics[width=9cm]{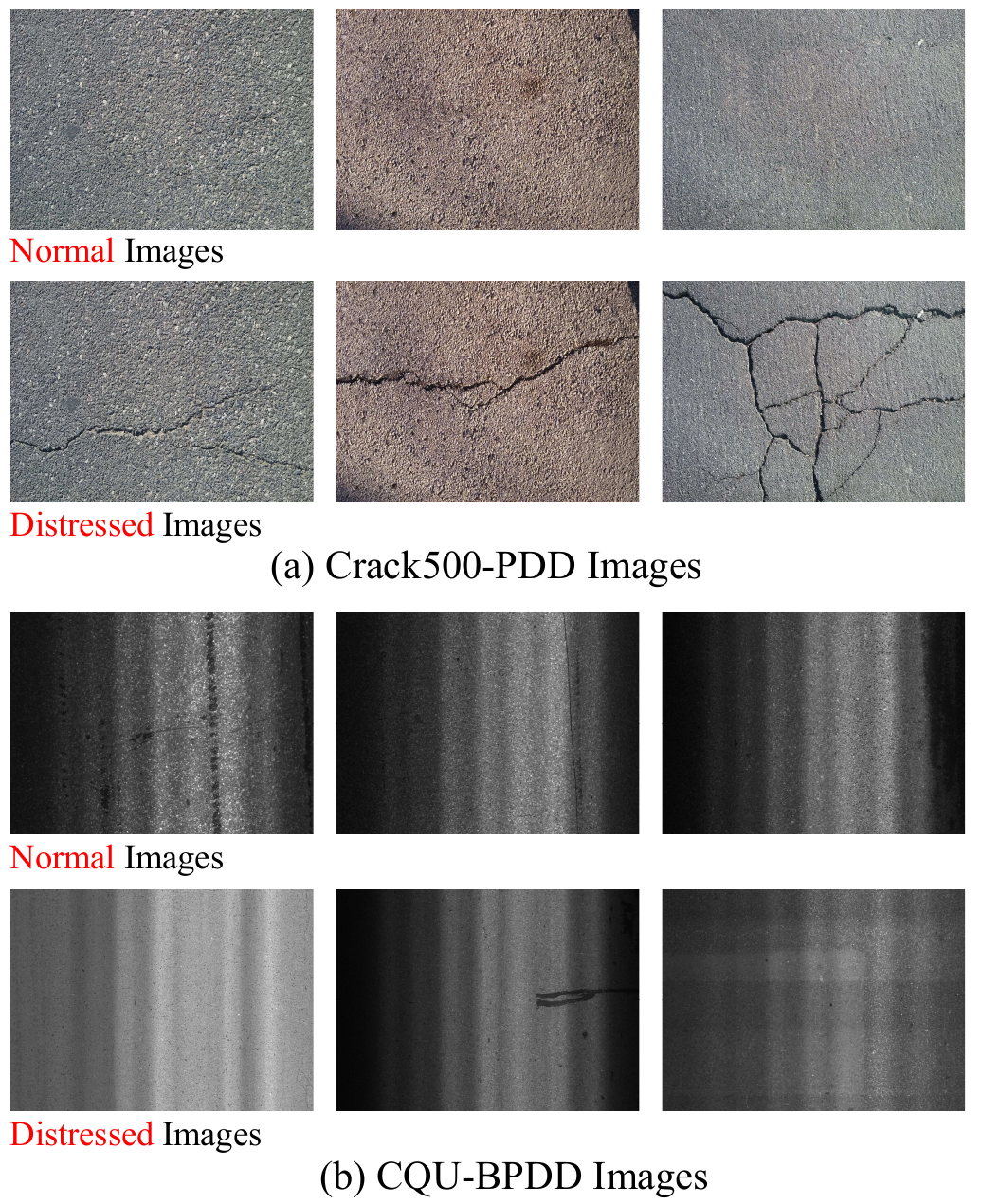}
    \caption{The examples of the augmented Crack500 (Crack500-PDD) dataset and the CQU-BPDD dataset. The CQU-BPDD dataset is a large-scale pavement image dataset, which consists of samples with different distresses not just limited to cracks, such as Longitudinal Crack, Cementation Fissure, and Mending. The Crack500-PDD dataset is a smaller-scale dataset constructed from the well-known crack detection dataset--Crack500. The Crack500 dataset is originally designed for crack detection or segmentation without containing any normal pavement images. On the contrary, the CQU-BPDD dataset is originally designed for pavement image classification.}\label{fig:crack500}
    \vspace{-0.4cm}
  \end{figure}

\subsubsection{Pavement Image Datasets}
We employed two pavement image datasets for evaluation. One is the CQU-BPDD dataset, and the other is the augmented Crack500 (Crack500 for Pavement Distress Detection, Crack500-PDD) dataset, which is a newly constructed pavement image classification dataset augmented from the well-known crack detection dataset--Crack500. The examples of these datasets are shown in Figure~\ref{fig:crack500}.

The CQU-BPDD is a large-scale bituminous pavement distress dataset, which is specially designed for pavement image classification and enables evaluating the approaches under all four application settings~\cite{c16}. This dataset involves seven different types of distress: alligator crack, crack pouring, longitudinal crack, massive crack, transverse crack, raveling, and mending. For settings of \textbf{I-DET}, \textbf{I-REC}, and \textbf{II-REC(n)}. We simply follow the data split strategy in \cite{c16}. With regard to the setting of \textbf{II-REC(i)}, 5140 distressed pavement images are randomly selected as the training set, while the rest 11,589 distressed pavement images are for testing. The detailed data split information of different settings is tabulated in Table~\ref{settings}.

In order to better assess our methods, we also have constructed a new pavement image dataset named Crack500-PDD via augmenting the Crack500 dataset~\cite{yang2019crack500} for Pavement Distress Detection, which is a well-known crack detection dataset. Since the Crack500 dataset only has distress pavement images, we produce some normal pavement images via erasing the crack areas of some carefully selected crack images with their corresponding crack segmentation annotations. Specifically, the normal ones are recovered from their distressed version via replacing the diseased pixels with their neighbor pixels based on the pixel-level labels provided by the dataset. The Crack500-PDD dataset is a small-scale pavement image dataset, which only contains 494 crack images and 286 normal pavement images. The Crack500-PDD dataset can only be leveraged for evaluating the pavement distress detection performances, since it only has normal and crack samples. In an experiment, we randomly select half of the samples for training, while the rest for validation. Because of the small scale of the dataset, we repeated this experiment five times and report the average performance to reduce the impacts of data partitioning.

\subsubsection{Implement Details}
Similar to IOPLIN, we adopt EfficientNet-B3 as the Patch Label Inference Network (PLIN). Since the comprehensive decision network (CDN) adopts a fully connected layer with fixed dimensions, WSPLIN requires the input size to be fixed at $300\times300$, and the optimizer uses RangerLars, which is just a combination of RAdam\cite{c24}, LookAhead\cite{c25} and LARS\cite{c26}. With regard to the Crack500-PDD, we first resized input images to 1200$\times$900, and converted them into gray images for a fair comparison. The learning rate is $8\times 10^{-4}$, and the cosine annealing strategy is adopted to adjust the learning rate: the learning rate remained unchanged in the first 25\% of the training process, and gradually decreased with the cosine function in the subsequent training process. Data augments such as rotation, flipping, and brightness balance are carried out for the raw images. The dropout rate of the classification layer is 0.5.

\begin{table}[tb]
    \small
    \caption{The detailed information of four different application settings on CQU-BPDD.} \label{settings}
    \small
    \begin{tabularx}{9cm}{p{1.4cm}p{1.8cm}<{\centering}p{0.9cm}<{\centering}p{0.6cm}<{\centering}p{0.9cm}<{\centering}p{0.6cm}<{\centering}}
        \toprule
        \multirow{2}{*}{Setting}& \multirow{2}{*}{Classifier}&\multicolumn{2}{c}{Train} & \multicolumn{2}{c}{Test} \\
        \cmidrule(r){3-4}\cmidrule(r){5-6}
        && \#Sample  & \#Class & \#Sample  & \#Class \\
        \midrule
        \textbf{I-DET} & Detector & 10137 & 2 & 49919 & 2\\[2pt]
        \hdashline[1.5pt/1.5pt]
        \textbf{I-REC} & I-Recognizer & 10137 & 8 & 49919 & 8 \Tstrut\\[2pt]
        \hdashline[1.5pt/1.5pt]
        \textbf{II-REC(i)} & II-Recognizer & 5140 & 7 & 11589 & 7  \Tstrut\\[2pt]
        \hdashline[1.5pt/1.5pt]
        \multirow{2}{*}{\textbf{II-REC(n)} } & Detector & 10137 & 2 & \multirow{2}{*}{49919} & \multirow{2}{*}{8} \Tstrut\\[1.5pt]
        &II-Recognizer& 5140 & 7 &  & \\[2pt]
        \bottomrule
        \end{tabularx}
    \vspace{-0.5cm}
\end{table}

\vspace{-0.2cm}
\subsection{Evaluation Metrics}
\subsubsection{Evaluation Metrics of Detection}
For pavement distress detection task, we adopt Area Under Curve (AUC) of Receiver Operating Characteristic (ROC)~\cite{auc}, which is common in binary classification tasks (this metric is not affected by classification threshold). It is mathematically defined as follows,
\begin{equation}\label{AUC}
    AUC=\frac{S_{p}-N_{p}\left ( N_{p}+1 \right )/2}{N_{p}N_{n}}
\end{equation}
where $S_{p}$ is the sum of all positive samples ranked, while $N_{p}$ and $N_{n}$ denote the number of positive and negative samples.
Additionally, Binary $F1$ score, which is the harmonic mean of precision and recall, is used to measure the models more comprehensively. The binary $F1$ score is defined as:
\begin{equation}\label{f1_bin}\small
    F1_{binary}= \frac{2\times \textbf{P}\times\textbf{R}}{\textbf{P} + \textbf{R}},\quad \textbf{P}=\frac{TP}{TP+FP},\quad\textbf{R}=\frac{TP}{TP+FN},
\end{equation}
where $\textbf{P}$ is the precision while \textbf{R} is the recall.  $TP$, $FP$, and $FN$ are the numbers of true positives, false positives and false negatives respectively. The precision measures how many true positive samples are among the samples that are predicted as positive samples. Similarly, recall measures how many true positive samples are correctly detected among all positive samples. Moreover, in the medical or pavement image analysis tasks, it is more meaningful to discuss the precision under the high recall, since the miss of the positive samples (the distressed sample) may lead to a more serious impact than the miss of the negative ones.
\subsubsection{Evaluation Metrics of Recognition}
For pavement distress recognition task, we mainly use the Top-1 accuracy and Marco $F1$ score to evaluate the performance of models. Top-1 accuracy mainly measures the overall accuracy of the models, while Marco $F1$ score evaluates the accuracy of the model across different categories. The Macro $F1$ score can be mathematically represented as follows,
\begin{equation}\label{f1_marco}
    F1_{marco}=\frac{1}{c}\sum_{i}^{c} F1_{binary}^{i},
\end{equation}
where $F1_{binary}^{i}$ indicates the binary $F1$ score of the $i$-th category, and $c$ is the total number of categories.

\textbf{Note:} The $F1$ represents $F1_{binary}$ and $F1_{marco}$ in pavement distress detection and recognition tasks respectively.

\vspace{-0.2cm}
\subsection{Baselines}
Histogram of Oriented Gradient (HOG)~\cite{c27}, Local Binary Pattern (LBP)~\cite{lbp}, Fisher Vector(FV)~\cite{fv}, Support Vector Machine (SVM)~\cite{c23},  ResNet-50~\cite{c19}, Inception-v3~\cite{c13}, VGG-19~\cite{c17}, ViT-S/16~\cite{vit}, ViT-B/16~\cite{vit}, EfficientNet-B3~\cite{c21}, Spatial Transformer Networks (STN)~\cite{jaderberg2015spatial,stn}, and Iterative Optimized Patch Label Inference Network (IOPLIN)~\cite{c16} are selected as baselines. The first four approaches are the shallow learning-based approaches. ResNet-50, Inception-v3, VGG-19, and EfficientNet-B3 are the classical Convolutional Neural Network (CNN) models. ViT-S/16 and ViT-B/16 are the recently popular transformer models. IOPLIN is a well elaborated pavement distress detection approach. The architectural design of STN is derived from~\cite{stn}.

\begin{table}[tb]
    \small
    \caption{The pavement distress detection performances of different methods on the augmented Crack500 (Crack500-PDD) dataset. We report the average performances over five random validations where each validation randomly retains half samples for training, while the rest for testing. Each result in this table is reported in \emph{ mean ($\pm$ std)}. In these experiments, all models were trained with ImageNet-1k pre-trained model and the SGD optimizer except ViT-B and ResNet-50. ViT-B is trained with the ImageNet-21k pre-trained model, while the optimizer of ResNet-50 is AdamW.}
  \begin{tabular}{lc<{\centering}c<{\centering}}
        \toprule
        Detectors(\textbf{I-DET})& AUC & P@R=95\%  \\
        \midrule
        EfficientNet-B3~\cite{c21} & 99.4\% ($\pm$ 0.25\%)  & 99.1\% ($\pm$ 0.7\%)\\[1.5pt]
        ViT-B (21k)~\cite{vit} & 98.9\% ($\pm$ 2.2\%) & 96.3\% ($\pm$ 7.3\%)\\[1.5pt]
        ResNet-50~\cite{c19} & 99.9\% ($\pm$ 0.2\%) & 100.0\% ($\pm$ 0.0\%)\\[1.5pt]
        VGG-19~\cite{c17} & 99.9\% ($\pm$ 0.1\%)  & 99.8\% ($\pm$ 0.2\%)\\[1.5pt]
        Inception-v3~\cite{c13} & 99.8\% ($\pm$ 0.1\%)  & 100.0\% ($\pm$ 0.0\%)\\[1.5pt]
        STN~\cite{stn} & 95.6\% ($\pm$ 2.7\%)  & 85.2\% ($\pm$ 7.4\%)\\[1.5pt]
        IOPLIN~\cite{c16} & 99.4\% ($\pm$ 0.4\%)  & 98.6\% ($\pm$ 1.0\%)\\[1.5pt]
        \textbf{WSPLIN-IP} & \textbf{100.0\% ($\pm$ 0.0\%)}  & \textbf{100.0\% ($\pm$ 0.0\%)}\\[1.5pt]
        \bottomrule
        \end{tabular}
    \label{crack500}
  \end{table}

\vspace{-0.4cm}

\begin{table}[tb]
    \small
    \caption{The pavement distress detection performances of different approaches on the CQU-BPDD detection benchmark. P@R = $n$\% indicates the precision when the corresponding recall is equal to $n$\%. }
    \centering
    \begin{tabularx}{9cm}{l p{0.6cm}<{\centering} X<{\centering}X<{\centering}p{0.6cm}<{\centering}}
    \toprule
     Detectors(\textbf{I-DET}) & AUC & P@R=90\% & P@R=95\% & $F1$\\
  \midrule
    HOG+PCA+SVM~\cite{c27}& 77.7\% &31.2\%&28.4\%& -\\[1.5pt]
    LBP+PCA+SVM~\cite{lbp}&82.4\% &34.9\%&30.3\%& -\\[1.5pt]
    HOG+FV+SVM~\cite{fv}&88.8\% &43.9\%&35.4\%& -\\[1.5pt]
    ResNet-50~\cite{c19}  & 90.5\% & 45.0\% & 35.3\% & -\\[1.5pt]
    Inception-v3~\cite{c13} & 93.3\% & 56.0\% & 42.3\% & -\\[1.5pt]
    VGG-19~\cite{c17}  & 94.2\% & 60.0\% & 45.0\% & -\\[1.5pt]
    ViT-S/16~\cite{vit}  & 95.4\% & 67.7\% & 51.0\% & 81.1\%\\[1.5pt]
    ViT-B/16~\cite{vit}  & 96.1\% & 71.2\% & 56.1\% & 80.6\%\\[1.5pt]
    EfficientNet-B3~\cite{c21} & 95.4\% & 68.9\% & 51.1\% & 81.3\%\\[1.5pt]
    STN~\cite{stn} & 94.9\% & 65.0\% & 48.8\% & 78.0\%\\[1.5pt]
    IOPLIN~\cite{c16} & 97.4\% & 81.7\% & 67.0\% & 85.3\%\\[1.5pt]
    {\bf WSPLIN-IP} & {\bf 97.5\%} & {\bf 83.2\%} & {\bf 69.5\%} & {\bf 86.4\%}\\[1.5pt]
     \midrule
    Recognizers(\textbf{I-REC}) & AUC & P@R=90\% & P@R=95\% & $F1$\\[2pt]
      \midrule
    EfficientNet-B3~\cite{c21}  & 96.0\% & 77.3\% & 59.9\% & 83.2\%\\[1.5pt]
    {\bf WSPLIN-IP} & {\bf 97.6\%} & {\bf 85.3\%} & {\bf 72.6\%} & {\bf87.4}\%\\[1.5pt]
    \bottomrule
    \end{tabularx}
    \label{tab:compare_binary}
    \vspace{-0.2cm}
\end{table}

\subsection{Pavement Distress Detection}
\subsubsection{Crack500-PDD}
Table~\ref{crack500} tabulates the pavement distress detection performances of different approaches on augmented Crack500 (Crack500-PDD). We only conduct evaluation under \textbf{I-DET}, since this dataset only contains two categories, namely crack and normal. From the observations, our method performs the best. However, we also observe that all methods actually achieve nearly perfect performances. For example, WSPLIN-IP, Inception-V3 and ResNet-50 all get 100\% accuracies in P@R=95\%, and the AUC of all methods are higher than 98\%. These phenomena imply that the pavement distress detection issue on this dataset is already well addressed, and this dataset is too simple to assess the performances of different pavement distress detection methods. We attribute this to three facts. The first one is the small scale of this dataset. The Crack500-PDD dataset only contains hundreds of samples. The second one is that the distress features of crack images are very salient, which is much easier to be distinguished from the normal ones, since this dataset is originally designed for crack detection, which focuses more on the extraction of complex crack structures. The third one is that this dataset lacks diversity in distress type. This dataset only contains crack images. However, the distress type is not limited to the cracks, which can be extended to even cracks that have different finer-grained types. In summary, it may not be unsuitable to employ the crack detection dataset for evaluating pavement image classification. Therefore, we further conducted a more systematic study on the CQU-BPDD dataset, which is a much larger dataset and specially designed for validating pavement distress detection and recognition performances.

\subsubsection{CQU-BPDD}
Table~\ref{tab:compare_binary} reports pavement distress detection performances of different approaches on CQU-BPDD. We evaluate our methods on all pavement distress detection settings. These approaches include the detectors trained under \textbf{I-DET} and the recognizers trained under \textbf{I-REC} where recognizers address the detection issue along with the recognition task. Based on observations, WSPLIN-IP outperforms all baselines in all evaluation metrics under both \textbf{I-DET} and \textbf{I-REC}. In \textbf{I-DET}, WSPLIN-IP improves the performances of IOPLIN by 0.1\%, 1.5\%, 2.5\%, and 1.1\% in AUC, P@R=90\%, P@R=95\%, and $F1$ respectively. In \textbf{I-REC}, WSPLIN-IP achieves 1.6\%, 8.0\%, 12.7\%, and 4.2\% performance gains over EfficientNet-B3 in AUC, P@R=90\%, P@R=95\%, and $F1$ respectively. Moreover, the methods under \textbf{I-REC} consistently perform much better than the ones under~\textbf{I-DET}. For example, the WSPLIN-IP trained under \textbf{I-REC} achieves 0.1\%, 2.1\%, 3.1\%, and 1.0\% performance gains than the WSPLIN-IP trained under \textbf{I-DET} in AUC, P@R=90\%, P@R=95\%, and $F1$ respectively. Similarly, the gains of EfficientNet-B3 are 0.6\%, 8.4\%, 8.8\%, and 1.9\%. We attribute this to the fact that recognizers trained under \textbf{I-REC} utilize fine-grained distress labels instead of binary distress labels for training the pavement image classification models. It reflects that the much finer-grained supervised information, such as the specific pavement distress information, can benefit pavement distress detection.

\begin{table}[tb]
    \small
    \centering
    \vspace{-0.3cm}
    \caption{The pavement distress recognition performances of different methods under different settings, namely \textbf{I-REC}, \textbf{II-REC(n)} and \textbf{II-REC(i)}. "Para." indicates the parameter scale of the deep learning model. Top-1 indicates the top-1 accuracy.}
    \begin{tabularx}{9cm}{p{3.3cm} X<{\centering}X<{\centering}X<{\centering} }
    \toprule
    Recognizers(\textbf{I-REC})  &Para. & Top-1& $F1$\\\midrule
ResNet-50 \cite{c19}&23M & 88.3\%  & 60.2\%    \\[1.5pt]
VGG-16 \cite{c17}&134M & 87.7\%  & 58.4\%      \\[1.5pt]
ViT-S/16~\cite{vit}&22M & 86.8\%  & 59.0\%    \\[1.5pt]
ViT-B/16~\cite{vit}&86M & 88.1\%  & 61.2\%    \\[1.5pt]
Inception-v3 \cite{c13}&22M & 89.3\%  & 62.9\%\\[1.5pt]
EfficientNet-B3~\cite{c21}&11M & 88.1\% &63.2\%\\[1.5pt]
STN~\cite{stn}&11M & 79.1\% &21.8\%\\[1.5pt]
\textbf{WSPLIN-IP}&\textbf{11M} & \textbf{91.1\%} &\textbf{66.3\%}\\[1.5pt] \midrule
    Recognizers(\textbf{II-REC(n)})  &Para. & Top-1& $F1$\\
    \midrule
        ResNet-50 \cite{c19}&23M & 82.8\%  & 53.6\%    \\[1.5pt]
    VGG-16 \cite{c17}&134M & 86.5\%  & 55.0\%      \\[1.5pt]
    ViT-S/16~\cite{vit}&22M & 86.7\%  & 56.6\%    \\[1.5pt]
    ViT-B/16~\cite{vit}&86M & 87.5\%  & 59.6\%    \\[1.5pt]
    Inception-v3 \cite{c13} &22M& 88.3\%  & 59.8\%     \\[1.5pt]
    EfficientNet-B3 \cite{c21}&11M & 88.9\%  & 61.2\%    \\[1.5pt]
    STN~\cite{stn}&11M & 76.8\% & 18.1\%     \\[1.5pt]
    \textbf{WSPLIN-IP}    & \textbf{11M}        & \textbf{90.0\%}  & \textbf{64.5\%}    \\[1.5pt]

    \midrule
     Recognizers(\textbf{II-REC(i)})  &Para. & Top-1& $F1$\\
    \midrule
    ResNet-50 \cite{c19}&23M & 71.2\%  & 61.5\%    \\[1.5pt]
    VGG-16 \cite{c17}&134M & 74.6\%  & 65.0\%      \\[1.5pt]
    ViT-S/16~\cite{vit}&22M & 75.0\%  & 64.9\%    \\[1.5pt]
    ViT-B/16~\cite{vit}&86M & 75.3\%  & 67.0\%    \\[1.5pt]
    Inception-v3 \cite{c13} &22M& 77.6\%  & 69.8\%     \\[1.5pt]
    EfficientNet-B3 \cite{c21}&11M & 78.6\%  & 70.3\%    \\[1.5pt]
    STN~\cite{stn}&11M & 65.1\% & 54.8\% \\[1.5pt]
    \textbf{WSPLIN-IP}    & \textbf{11M}        & \textbf{85.0\%}  & \textbf{77.2\%}    \\[1.5pt]
    \bottomrule
    \end{tabularx}
    \vspace{-0.5cm}
\label{rec}
\end{table}

\vspace{-0.2cm}
\subsection{Pavement Distress Recognition}
Table~\ref{rec} records the pavement distress recognition performances and parameter scales of different approaches under different application settings on the CQU-BPDD dataset. Similar to the pavement distress detection performances, WSPLIN-IP achieves better recognition performances than baselines under all settings, while enjoys a smaller parameter scale. In \textbf{I-REC}, the performance gains of WSPLIN-IP over Inception-v3, which is the second-best method, are 1.8\% and 3.4\% in top-1 accuracy and $F1$, respectively. In~\textbf{II-REC(n)} and \textbf{II-REC(i)}, the EfficientNet-B3 achieves the second-best performance. The performance gains of WSPLIN-IP over it under \textbf{II-REC(n)} are 1.1\% and 3.3\% in top-1 accuracy and $F1$, respectively. Such gains under \textbf{II-REC(i)} are 6.4\% and 6.9\%. The distribution of different pavement image categories is imbalanced. Top-1 accuracy is sensitive to this data imbalance, while $F1$ is more stable to this imbalance. Therefore, $F1$ can better reflect the comprehensive performances of recognizers. According to the observations, WSPLIN-IP shows more advantages compared with baselines in $F1$.

The test settings of \textbf{I-REC} and \textbf{II-REC(n)} are identical as seen in Table~\ref{settings}. However, the models trained under~\textbf{I-REC} outperform the ones of \textbf{II-REC(n)}. For example, Inception-v3, ViT-B/16, EfficientNet-B3, and WSPLIN-IP trained under \textbf{I-REC} achieve 3.1\%, 1.6\%, 2.0\%, and 1.8\% improvements over the ones under \textbf{II-REC(n)} in $F1$. This implies that the end-to-end pavement distress recognition solution, which addresses detection and recognition tasks jointly (\textbf{I-REC}), enjoys more advantages than the conventional two-stage implementation solution, which addresses detection and recognition tasks individually (\textbf{II-REC(n)}), in the real-world application. We attribute this to that the end-to-end solution exploits the complementarity of these two tasks and introduces global optimization.

An interesting phenomenon is observed from Table~\ref{rec} that the top-1 accuracies of \textbf{II-REC(i)} are lower than the ones of the rest two settings while its F1-scores are higher than the F1-scores of the rest settings. This is because the test setting of \textbf{II-REC(i)} is different from the settings of \textbf{I-REC} and \textbf{II-REC(n)}, which does not involve any normal pavement sample. Top-1 accuracy is measured in sample-wise but $F1$ is measured in category-wise. The normal samples comprise a large proportion of the whole data in \textbf{I-REC} and \textbf{II-REC(n)}. Therefore, the superabundant normal samples will make the recognizers trained under \textbf{I-REC} bias to the classification of the normal sample, which leads to the high top-1 accuracy but the low $F1$. With regard to \textbf{II-REC(n)}, the massive normal samples push up the top-1 accuracy. However, the measure of $F1$ is independent of the sample amount and the classification error of \textbf{II-REC(n)} is accumulated from both the detection and recognition stages. Therefore, it achieves a lower $F1$ in comparison with \textbf{II-REC(i)}.

\vspace{-0.2cm}

\begin{table}[tb]
    \vspace{-0.3cm}
        \small
        \caption{The ablation study results of the proposed approaches under the same memory in different settings and evaluation metrics. (w/o indicates without and PLSC is the patch label sparsity constraint.)}
     \begin{tabular}{p{3.15cm}p{1.3cm}<{\centering}p{1.2cm}<{\centering}c}
            \toprule
            \textbf{Method} & \tabincell{c}{\textbf{I-DET}\\(P@R=90\%)} & \tabincell{c}{\textbf{I-REC} \\($F1$)}& \tabincell{c}{\textbf{TrainTime}\\(Reduction)}\\
            \midrule
            IOPLIN &  81.7\% &  - &12.5h\\
            WSPLIN-SW &  80.9\% &  64.4\% &9.6h (-23\%)\\[1.5pt]
            WSPLIN-IP w/o PLSC & 81.2\%&64.5\% &11.0h (-12\%) \\[1.5pt]
            WSPLIN-IP &  \textbf{83.2\%} &  \textbf{66.3\%} &11.1h (-11\%)\\[1.5pt]
            WSPLIN-SS ($\alpha = 0.25$) &  81.1\% &  64.1\%&  \textbf{3.2h (-74\%)}\\[1.5pt]
           WSPLIN-SS  ($\alpha = 0.50$) &  81.4\% & 64.9\% &5.7h (-54\%)\\[1.5pt]
            WSPLIN-SS  ($\alpha = 0.75$) &  80.0\% &  63.7\% & 8.4h (-33\%)\\[1.5pt]
            \bottomrule
            \end{tabular}
        \label{ablation}
        \vspace{-0.4cm}
    \end{table}

\subsection{Ablation Study}
In this section, we evaluated the impact of various components and hyperparameters on the performance and efficiency of our model, as well as compared to IOPLIN. The results of these experiments are recorded in Table~\ref{ablation}.

\subsubsection{Discussion on Patch Collection Strategies}
We adopted three strategies named Slide Window (SW), Image Pyramid (IP), and Sparse Sampling (SS) to collect the patches from pavement images. Their corresponding versions are WSPLIN-SW, WSPLIN-IP, and WSPLIN-SS, respectively. In all three versions, WSPLIN-IP, which is the default version of WSPLIN, achieves the best performances under two application settings with different evaluation metrics. WSPLIN-IP achieves 1.5\% performance gains in P@R=90\% in the pavement distress detection case. However, its training time is only 89\% of the training time of IOPLIN. In comparison with WSPLIN-SW, WSPLIN-IP exploits not only the local information but also the scale information of pavement images. The results indicate that such scale information can further improve the performance of WSPLIN. Although WSPLIN-SW has not outperformed IOPLIN with a 0.8\% performance decrease in pavement distress detection, WSPLIN-SW is much faster than WSPLIN-IP and its training only takes around 3/4 of the training time of IOPLIN. We attribute this to the efficiency advantage of the end-to-end model optimization strategy. Moreover, compared with other versions, WSPLIN-SW has not suffered from the scale variation, so it can produce better patch label inference visualization results and thereby enjoys better interpretability.

WSPLIN-SS also takes the scale information into consideration, and can be deemed as a lightweight version of WSPLIN-IP. The best-performed WSPLIN-SS ($\alpha = 0.5$) achieves similar performance as IOPLIN where $\alpha$ is a hyperparameter to control the number of sampled patches in each layer of an image pyramid. However, WSPLIN-SS saves around half of the training time compared with IOPLIN, and it is only 76\% of the training time of WSPLIN-IP. Clearly, WSPLIN-SS highly speeds up WSPLIN only with an acceptable performance decrease. Another interesting phenomenon is observed that WSPLIN-SS with a higher $\alpha$ does not always enjoy better performance. Generally, a higher $\alpha$ implies collecting more patches which means more information can be preserved for classification. However, the results indicate that not all preserved information is necessary for classification. Moreover, the less amount of patches per image means more images can be taken into one batch for model optimization since the memory size is fixed in our case. The higher diversity of pavement images in each batch benefits model optimization. A good sparse sampling strategy should optimize the trade-off between patch preservation and the diversity of samples in the same batch. We believe this is the reason why WSPLIN-SS ($\alpha = 0.50$) performs well in both tasks.

In summary, all WSPLIN approaches show prominent advantages in training efficiency with similar or even better performances. We recommend using WSPLIN-IP in the application scenarios, which is the best-performing version. And the lightweight WSPLIN-SS introduces a good trade-off between performance and efficiency benefits from its unique design.
In particular, WSPLIN-SW should be the ideal choice if the distress does not suffer from the high diversity in scale.
\subsubsection{Discussion on Patch Label Sparsity Constraint}
\begin{figure}[tb]
     \begin{center}
    \includegraphics[width=8.6cm]{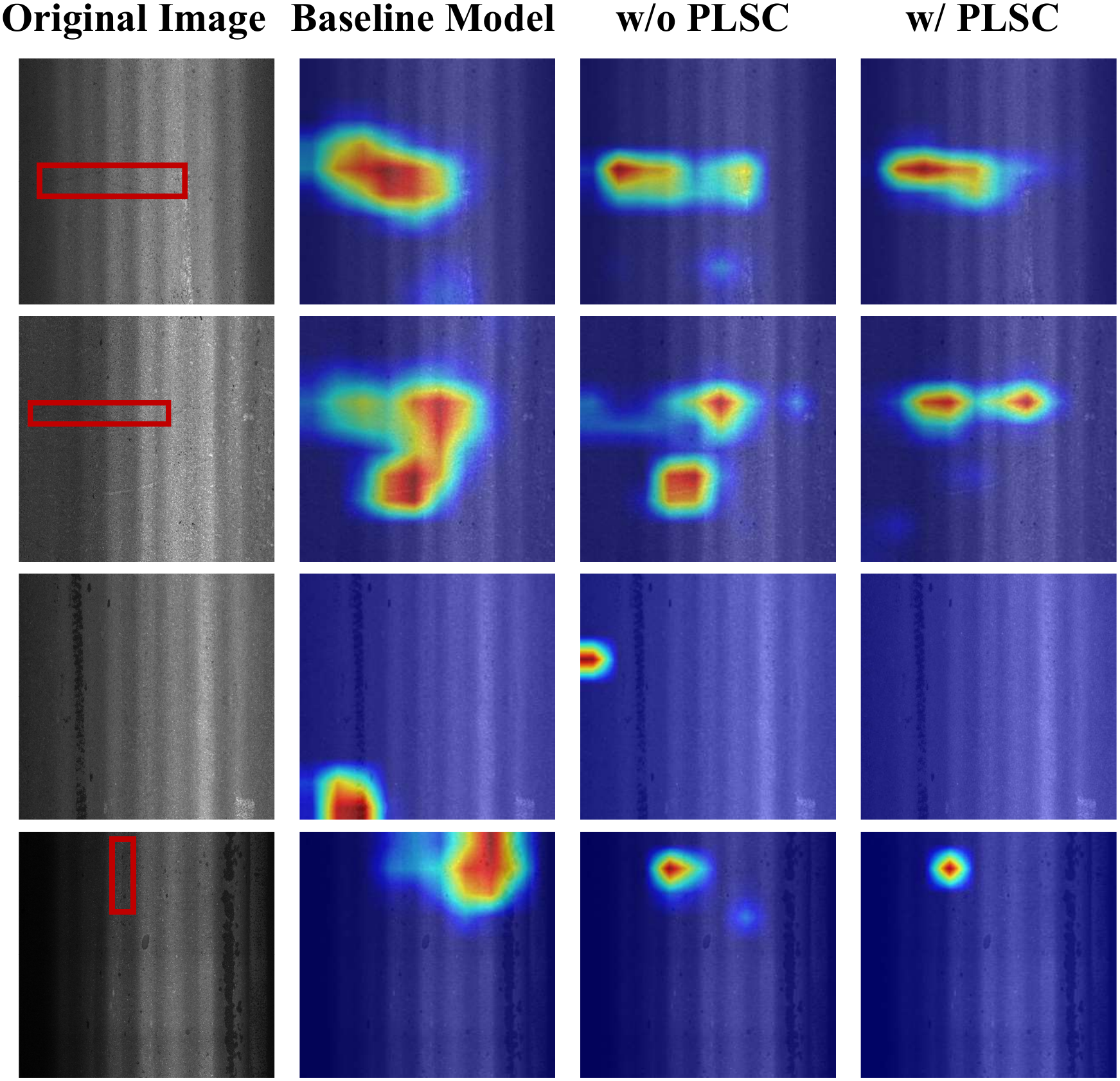}
    \caption{The Class Activation Map (CAM) visualizations of WSPLIN-IP models trained with or without PLSC. From the left column to the right column respectively indicate the ground truth, CAMs of baseline, CAMs of WSPLIN-IP without PLSC, and CAMs of WSPLIN-IP with PLSC.}
    \label{pic_heatmap}
    \vspace{-0.4cm}
    \end{center}
\end{figure}

The distressed area is often a small proportion of the whole image. Therefore, we introduce the Patch Label Sparsity Constraint (PLSC) to model and leverage this prior property for improving the discriminating power of the model and better addressing the pavement image classification issue. Table~\ref{ablation} reports the performances of WSPLIN-IP with and without PLSC. WSPLIN-IP with PLSC achieves 2.0\% more accuracies in P@R=90\% under \textbf{I-DET} and 1.8\% greater F1-scores under \textbf{I-REC} over WSPLIN without PLSC. This implies that PLSC can offer a considerable improvement to WSPLIN. We also leverage Grad-CAM~\cite{grad_cam} to plot the Class Activation Maps (CAM) of the features extracted by the WSPLIN-IP models before and after using PLSC in Figure~\ref{pic_heatmap}. The CAM visualization results also validate that PLSC benefits the distressed feature extraction.

\begin{figure}[tb]
    \begin{center}
    \includegraphics[scale=0.55]{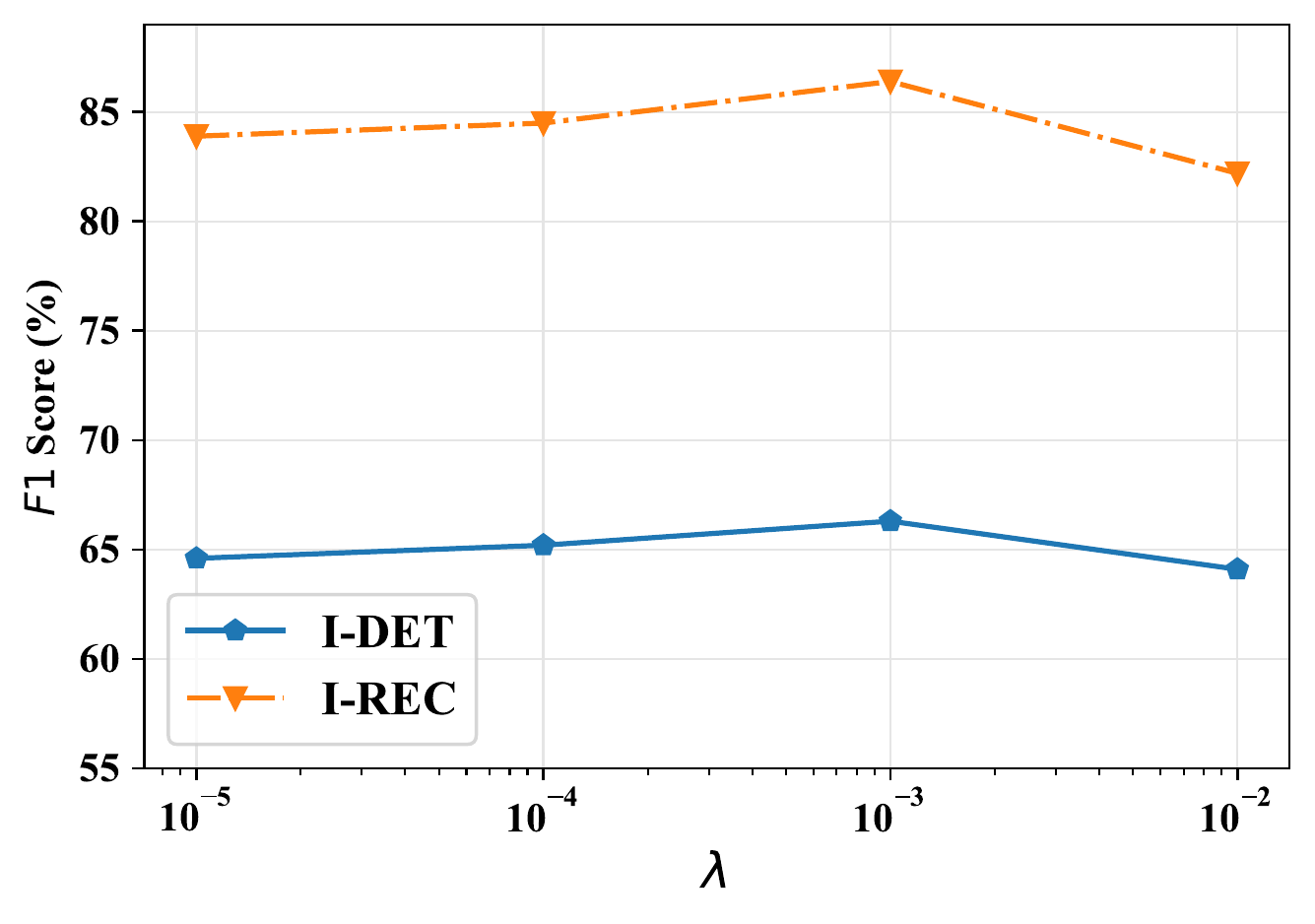}
    \vspace{-0.2cm}
        \caption{ The impacts of different $\lambda$ to the performance of WSPLIN-IP under \textbf{I-DET} and \textbf{I-REC}.}
           \label{pic_lambda}
    \end{center}
    \vspace{-0.5cm}
\end{figure}

$\lambda$ is a positive hyperparameter for reconciling the classification loss and the PLSC. Figure~\ref{pic_lambda} plots the relationships between the different values of $\lambda$ and the performances of WSPLIN-IP under \textbf{I-DET} and \textbf{I-REC}. From observations, we can find that the WSPLIN-IP is insensitive to the setting of $\lambda$. The optimal $\lambda$ is $10^{-3}$.

\subsubsection{The Efficiency of WSPLIN}
According to observations in Table~\ref{ablation}, all WSPLIN approaches are more efficient than IOPLIN. Moreover, IOPLIN and WSPLIN have very similar network structure, so they have the same parameter scale.

\vspace{-0.2cm}
\subsection{Weakly and Self Supervised Analysis}
\begin{table}[tb]
    \small
    \centering
    \caption{The pavement image classification performances of the proposed approach in comparison with some SOTA weakly and self-supervised learning approaches. Grad-CAM and C$^2$AM are Weakly Supervised Object detection (WSOD) approaches, and C$^2$AM can also be applied to address Weakly Supervised Semantic Segmentation (WSSS) issue. DINO and SimSiam are two influential Self-Supervised Image Classification (SSIC) approaches. Please note, our method and IOPLIN are Fully Supervised Image Classification (FSIC) approaches. Only the patch label inference parts of our method and IOPLIN are in a weakly supervised fashion.}
  \begin{tabular}{cc<{\centering}c<{\centering}c<{\centering}}
        \toprule
        \textbf{Method} & \textbf{Settings}& \tabincell{c}{\textbf{I-DET}\\(P@R=90\%)} & \tabincell{c}{\textbf{I-REC} \\($F1$)}\\
        \midrule
        Grad-CAM~\cite{selvaraju2017grad_cam} & WSOD& 68.9\% & 63.2\% \\[1.5pt]
        C$^2$AM~\cite{xie2022c2am} & WSOD$\setminus$WSSS& 59.0\% & 51.0\% \\[1.5pt]
        DINO~\cite{caron2021dino} & SSIC & 24.8\% & 11.0\% \\[1.5pt]
        SimSiam~\cite{chen2021simsiam} & SSIC & 24.8\% & 11.4\% \\[1.5pt]
        IOPLIN~\cite{c16} & FSIC& 81.7\% &  - \\[1.5pt]
        WSPLIN-IP & FSIC &  \textbf{83.2\%} &  \textbf{66.3\%} \\[1.5pt]
        \bottomrule
        \end{tabular}
    \label{wsl}
    \vspace{-0.6cm}
  \end{table}

Our method is a fully supervised learning approach from the perspective of image classification, since the output and input of our model are pavement images and category labels, respectively, and all this information is available for training. Since our method is a fully supervised learning method, why do we call it a Weakly Supervised Patch Label Inference Network (WSPLIN)? This is because we transform this image-level fully supervised learning problem as a patch-level weakly supervised learning issue for a solution. The main idea of our work is to divide images into patches and then use neural networks to infer the patch labels. Finally, the patch labels are aggregated to infer the label of images. The core step of our method is patch label inference. Since this step has no patch-level supervised information, the patch label inference is conducted in a weakly supervised learning manner.

In order to further analyze pavement image classification in different learning manners, we employ some recent advanced weakly and self-supervised learning approaches, namely Grad-CAM~\cite{selvaraju2017grad_cam}, C$^2$AM~\cite{xie2022c2am}, DINO~\cite{caron2021dino}, and SimSiam~\cite{chen2021simsiam}, for addressing pavement image classification issue. Grad-CAM and C$^2$AM are the weakly supervised learning approaches, while DINO and SimSiam are the self-supervised learning approaches.
The experimental results are reported in Table~\ref{wsl}. The results show that the fully supervised learning manner performs much better than the weakly and self-supervised learning manners. For example, the performance gains of WSPLIN-IP over Grad-CAM, C$^2$AM, DINO and SimSiam in F1 are 3.1\%, 15.3\%, 55.3\%, and 54.9\%, respectively. Grad-CAM is the best-performed weakly supervised learning approach. It is a post-processing method. Thus, it actually has not introduced any impact on the training process of the original classification model, and the classification performance only depends on its backbone--EfficientNet-B3. C$^2$AM is a SOTA one-stage weakly supervised learning method. However, it performs even worse than its backbone (-9.9\% on I-DET, -12.2\% on I-REC). We believe this is due to the interference from the pseudo label generation process, since all the processes are optimized together and the model is originally designed for object detection and segmentation instead of classification. The self-supervised learning approaches, such as DINO and SimSiam, almost fails in pavement distress detection and recognition tasks. This reveals the importance of supervised information in the feature learning step. Note, we here only use self-supervised learning methods to train the feature learning part and train the classifiers with supervised information. In summary, directly applying the weakly or self-supervised learning approaches to address pavement distress detection and recognition issues will lead to unsatisfactory performances, and the fully supervised fashion is still the best way for pavement image classification.

\begin{figure}[tb]
    \centering
    \includegraphics[width=9cm, keepaspectratio]{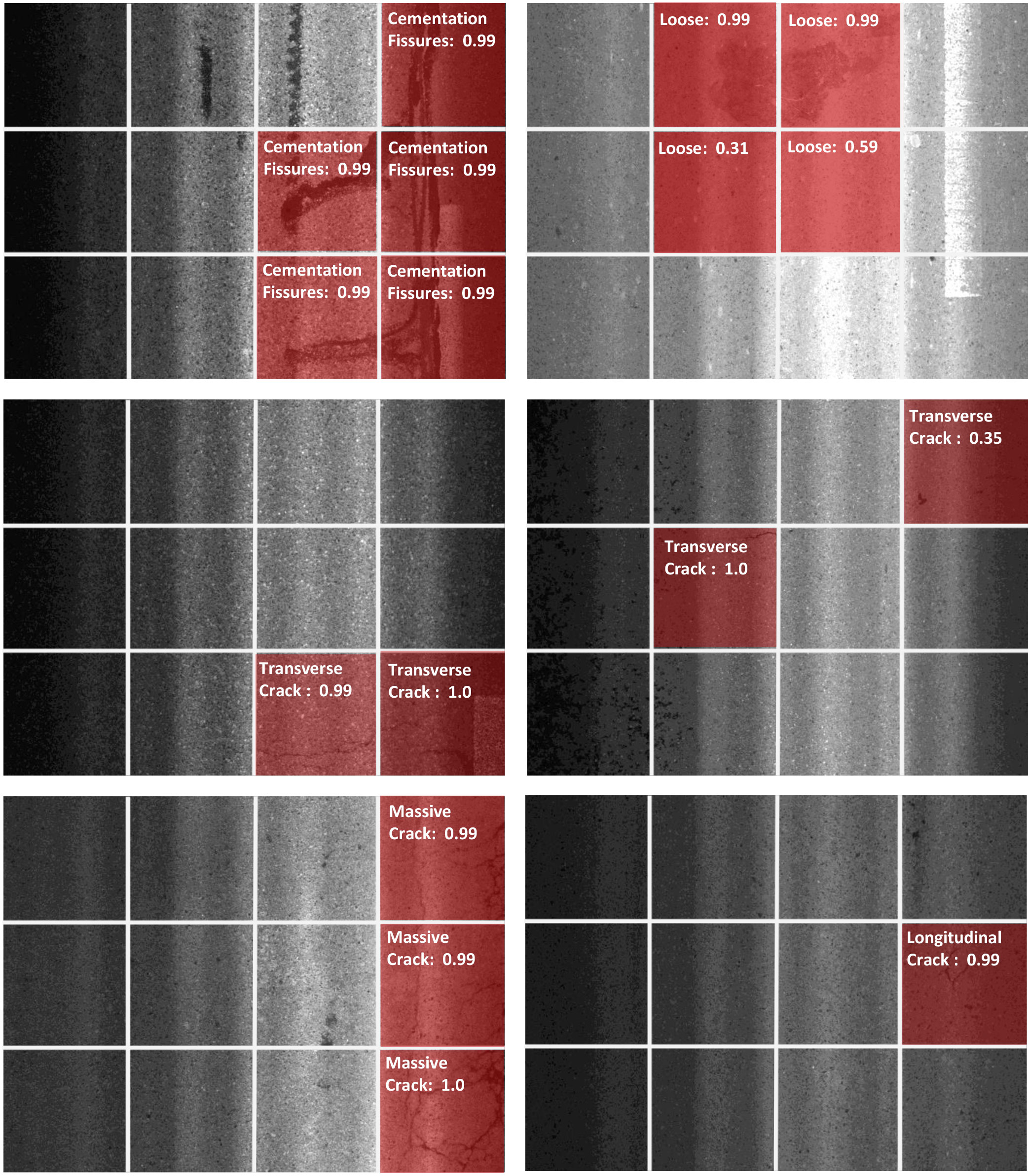}
    \caption{The visualizations of pavement images labeled by WSPLIN in the patch level. Best viewed in scale.}
    \label{fig:visualization}
    \vspace{-0.5cm}
\end{figure}

\vspace{-0.2cm}
\subsection{User Scenarios}
In highway maintenance, the engineer will use a professional pavement vehicle to capture pavement images of the highway at regular intervals. Once they obtain the pavement images, they will find out the distressed pavements, assess the severity of the distress, and then accomplish the pavement distress statistics for a highway. The pavement distress detection and recognition all have actual application values in this pipeline. For a 120-kilometer length highway with four lanes, each inspection will produce more than 240,000 pavement images. More than 95\% are actually normal pavement images that will not be further analyzed in the next step. Conventionally, the engineers must manually filter out these normal pavement images, leading to heavy labor costs. Like IOPLIN, WSPLIN enables engineers to filter out most normal images automatically, thereby significantly reducing manual labor costs. More specifically, WSPLIN often produces a confidence score for each pavement image to measure the probability that the image belongs to the diseased one. Thus, we can filter out most of the normal pavement images by setting a confidence score threshold.

Moreover, WSPLIN has wider application scenarios in comparison to IOPLIN. IOPLIN can only address the pavement distress detection problem, which is a typical binary image classification issue and attempts to find the distressed samples only. WSPLIN can tackle both the pavement distress detection and the recognition tasks under various aforementioned application settings shown in Table~\ref{settings}. Roughly distress localization and quality pavement distress recognition can assist engineers in automatically accomplishing detailed pavement distress statistics for roads over a while. This statistics information is crucial to assess the health situations of roads and speculate on the risk elements that cause pavement distress. This information is also vital for designing a suitable pavement maintenance strategy and estimating pavement maintenance expenditure. Moreover, not all pavement distresses are cracks, and some pavement distresses may need some special analysis. The pavement distress recognition system can automatically group the distress pavement images based on their distress types, thereby facilitating the engineers to analyze the different types of pavement distress further. Figure~\ref{fig:visualization} gives some examples of this scenario via visualizing the patch labels produced by the trained WSPLIN.

\vspace{-0.2cm}
\section{Conclusions}
\label{sec:conclusions}
In this paper, we present a novel patch-based deep learning model named WSPLIN for automatic pavement distress detection and recognition in the wild. WSPLIN divides the pavement image into patches with different patch collection strategies and then learns the label of patches in a weakly supervised manner. Finally, these inferred patch labels are fed into a comprehensive decision network for yielding the final recognition results. Similar to IOPLIN, WSPLIN can sufficiently utilize the resolution and scale information, and can also provide interpretable information, such as the location of the distressed area. However, WSPLIN is more efficient than IOPLIN with similar or even better performance. The experiments on a large pavement distress dataset validate the effectiveness of our approach. In the future, incorporating Spatial Transformer Networks (STN)~\cite{stn}, which is a popular technique in defect classification, into our method should be an interesting direction. Moreover, employing the self-supervised learning methods to pre-train our models may potentially further boost performances.

\footnotesize
\bibliographystyle{IEEEtran}
\bibliography{references}

%
\vspace{-1.2cm}
\begin{IEEEbiography}[{\includegraphics[width=1in,height=1.25in,clip,keepaspectratio]{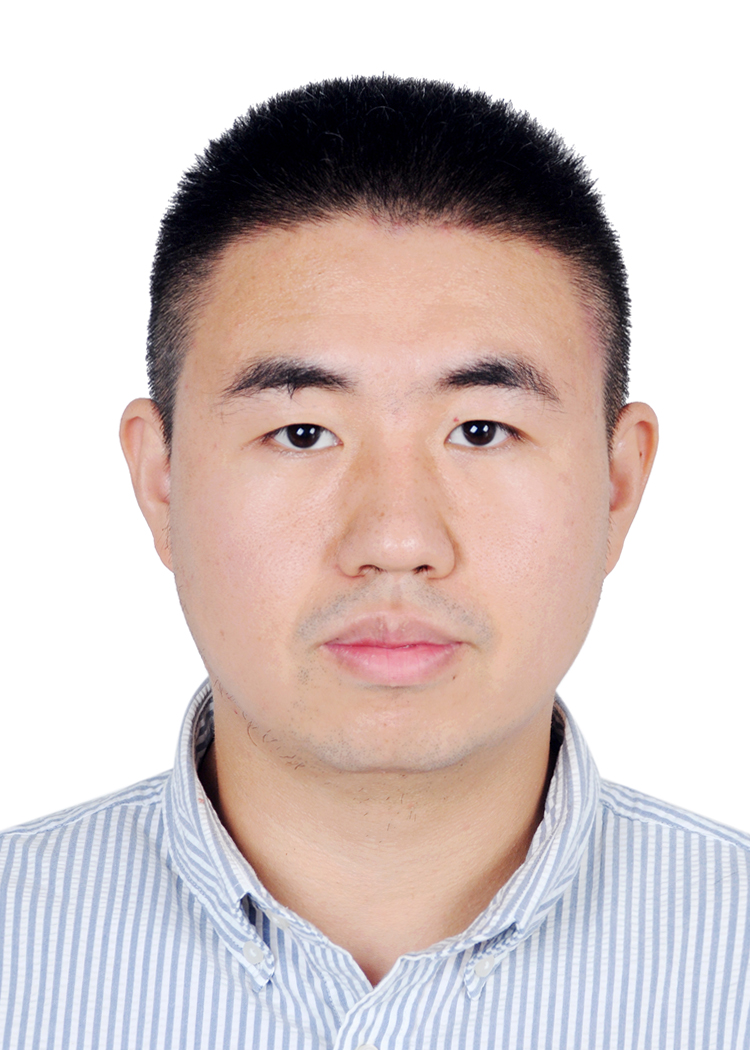}}]{Sheng Huang} (M'15)
received his BEng and PhD degrees both from Chongqing University, Chongqing, P.R.China, in 2010 and 2015 respectively. He was a visiting PhD student at the department of computer science, Rutgers University, New Brunswick, NJ, USA, from 2012 to 2014. He is currently an associate professor at the school of big data and software engineering, Chongqing University. He has authored/coauthored more than 30 scientific papers in venues, such as CVPR, AAAI, TIP, TIFS and TCSVT. His research interests include computer vision, machine learning, image processing and artificial intelligent applications.
\end{IEEEbiography}
\vspace{-1.3cm}
\begin{IEEEbiography}[{\includegraphics[width=1in,height=1.25in,clip,keepaspectratio]{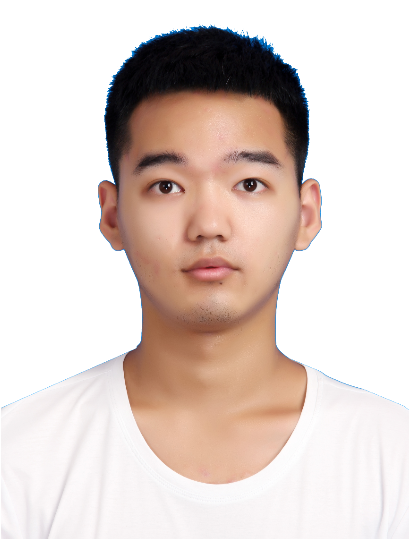}}]{Wenhao Tang}
is currently a master student in Big Data \& Software Engineering from Chongqing University (CQU), Chongqing, P.R. China. His research interests include computer vision, intelligent transportation systems, and artificial intelligent applications.
\end{IEEEbiography}
\vspace{-1.1cm}
\begin{IEEEbiography}[{\includegraphics[width=1in,height=1.25in,clip,keepaspectratio]{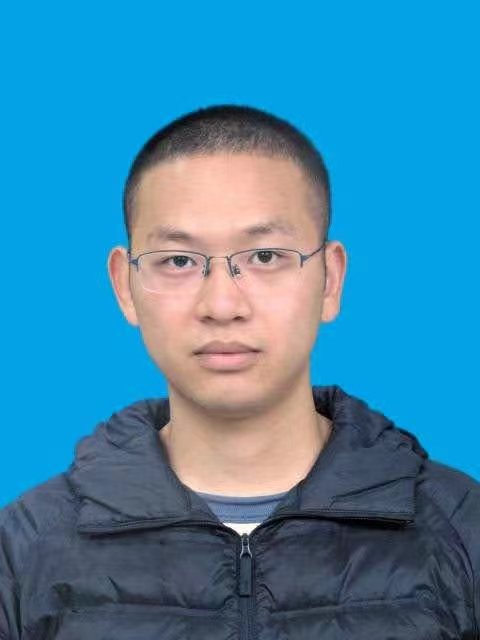}}]{Guixin Huang} (M'15)
is currently working at Global Function and Technology department of Citigroup as a backend developer. He obtained his Master's degree in software engineering from Chongqing University in 2021. His research fields mainly focus on image recognition and processing.
\end{IEEEbiography}
\vspace{-1.1cm}
\begin{IEEEbiography}[{\includegraphics[width=1in,height=1.25in,clip,keepaspectratio]{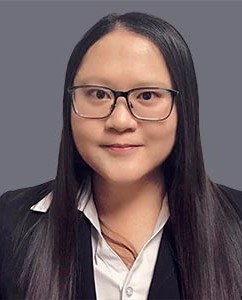}}]{Luwen Huangfu}
received the B.S. degree in software engineering from Chongqing University, Chongqing, China, the M.S. degree in computer science from the Chinese Academy of Sciences, Beijing, China, and the Ph.D. degree in management information systems from University of Arizona, Tucson, AZ, USA. She is currently an Assistant Professor with Fowler College of Business, San Diego State University, San Diego, CA, USA, where she is also with the Center for Human Dynamics in the Mobile Age. She has authored/coauthored more than 30 scientific papers in venues, such as IEEE Transactions on Cybernetics, IJCAI, ACM MM, Journal of Medical Internet Research (JMIR), IEEE Transactions on Intelligent Transportation Systems, Pacific Asia Journal of the Association for Information Systems, IEEE Signal Processing Letters, LREC, ICASSP, and IEEE ISI. Her research interests include business analytics, text mining, data mining, image processing, computer vision, artificial intelligence, and healthcare management.
\end{IEEEbiography}
\vspace{-1.1cm}
\begin{IEEEbiography}[{\includegraphics[width=1in,height=1.25in,clip,keepaspectratio]{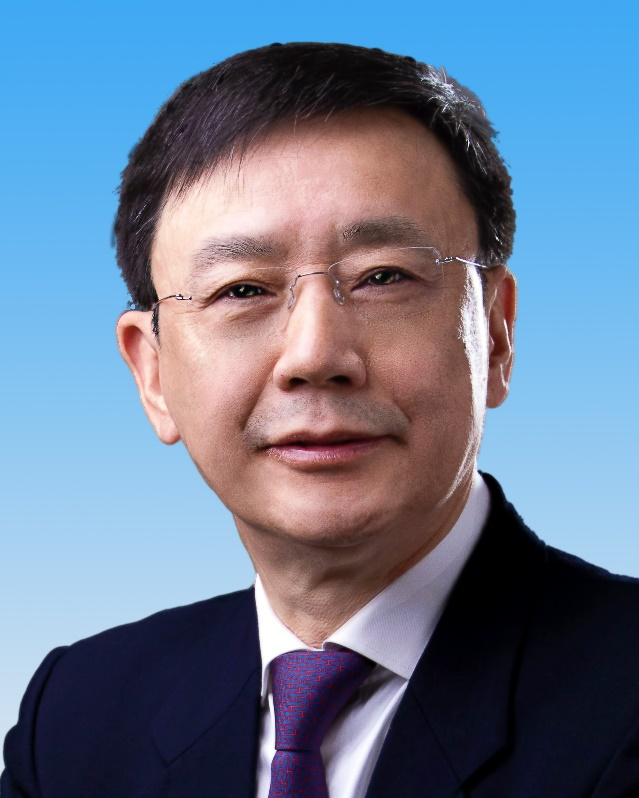}}]{Dan Yang}
received the B.Eng. degree in automation, the M.S. degree in applied mathematics, and the Ph.D. degree in machinery manufacturing and automation from Chongqing University, Chongqing. From 1997 to 1999, he held a post-doctoral position with the University of Electro-Communications, Tokyo, Japan. He is currently the President with Southwest Jiaotong University, and still holds the academic position with the School of Big Data and Software Engineering, Chongqing University. He has authored over 150 scientific papers and some of them are published in some authoritative journals and conferences, such as IEEE TRANSACTIONS ON PATTERN ANALYSIS AND MACHINE INTELLIGENCE, IEEE TRANSACTIONS ON IMAGE PROCESSING, IEEE TRANSACTIONS ON MEDICAL IMAGING, CVPR, and BMVC. His research interests include computer vision, image processing, pattern recognition, software engineering, and scientific computing.
\end{IEEEbiography}

%
%
%




\end{document}